\documentclass[11pt,twocolumn]{article}
\usepackage[utf8]{inputenc}
\usepackage[T1]{fontenc}
\usepackage{times}
\usepackage{geometry}
\usepackage{graphicx}
\usepackage{amsmath}
\usepackage{booktabs}
\usepackage{hyperref}
\usepackage{tabularx}
\usepackage{caption}
\usepackage{enumitem}
\usepackage{cite}
\usepackage{url}
\usepackage{xcolor}

\setlist{nosep, leftmargin=1.5em}
\hypersetup{colorlinks=true, linkcolor=blue, citecolor=blue, urlcolor=blue}

\title{\Large\textbf{Governing at Machine Speed: An Adaptive Intelligence\\Architecture for Real-Time AI Policy Enforcement}}

\author{
Sandeep Bokkasam\\[4pt]
\small\texttt{sandeep.b2013@yahoo.com}\\[10pt]
B. Durgalakshmi\\[4pt]
\small\texttt{durgalakshmi.cse@tagoreiet.ac.in}\\[8pt]
\small\textit{September 2026}
}

\date{}

\begin{document}
\maketitle

\begin{abstract}
Enterprise AI adoption has reached 78\% of organizations globally, yet the infrastructure to govern that adoption has not kept pace. This paper identifies and characterizes the \textit{attestation deficit}, a structural condition in which organizations maintain governance policies but cannot produce auditable, tamper-evident evidence of enforcement within regulatory timelines. Drawing on empirical data from the Stanford 2026 AI Index Report (362 documented incidents), the IBM/Ponemon 2026 Cost of a Data Breach study (USD 4.99M average cost, 92\% lacking access controls), and the EY/AIUC-1 Consortium survey (38\% end-to-end monitoring, 17\% agent-to-agent coverage), this paper demonstrates that the governance failure is organizational and architectural rather than technical. To address this deficit, we propose AGIL (Adaptive Governance Intelligence Layer), a conceptual five-layer architecture designed to use machine learning for real-time AI governance enforcement. The proposed layers include: (1) Autonomous Discovery for shadow AI detection via behavioral fingerprinting, (2) Behavioral Risk Classification unifying security, hallucination, privacy, and accountability scoring, (3) a Policy Enforcement Gateway for inline permit/deny/modify decisions at sub-100ms latency, (4) a Continuous Attestation Engine generating tamper-evident audit trails as a byproduct of enforcement, and (5) Adaptive Policy Intelligence for ML-driven policy evolution across jurisdictions. AGIL is presented as a theoretical framework and architectural proposal; empirical validation through controlled deployment remains a direction for future work. We conduct a design-level validation against three documented incidents from 2025 to 2026 and propose a phased deployment roadmap with measurable success metrics.
\end{abstract}


\section{Introduction}

The trajectory of enterprise AI adoption has outpaced governance infrastructure by a widening margin. Organizations have invested heavily in governance \textit{documentation}, including policies, frameworks, ethics boards, and vendor assessments, while building almost no infrastructure for governance \textit{enforcement}. The result is what this paper terms the \textit{attestation deficit}: a structural condition where governance exists on paper but cannot produce proof under regulatory scrutiny.

The empirical evidence supporting this characterization is substantial. Stanford's 2026 AI Index Report documented 362 AI-related incidents in 2025, a 56.4\% increase from 233 the year before~\cite{stanford2026}. IBM's 2026 Cost of a Data Breach Report, conducted by the Ponemon Institute across 602 organizations in 17 industries and 16 countries, found that global breach costs reached a record USD 4.99 million, with AI-related breaches averaging USD 5.33 million~\cite{ibm2026}. Among breached organizations, shadow AI incidents more than doubled from 20\% to 43\%, and 92\% of those suffering AI-related breaches lacked proper access controls. The report's most consequential finding was that 68\% of breached organizations had no AI governance policy at all, and of the six governance controls measured in both 2025 and 2026, five \textit{lost} adoption~\cite{ibm2026}.

The underlying problem is architectural rather than attitudinal. Existing governance frameworks, including the NIST AI Risk Management Framework~\cite{nist2023}, ISO/IEC 42001~\cite{iso42001}, and the EU AI Act~\cite{euaiact}, operate exclusively at the policy layer. They define what organizations \textit{should} do but provide no runtime enforcement machinery. When CrowdStrike's 2026 Global Threat Report documents AI-enabled lateral movement in 27 seconds~\cite{crowdstrike2026} and Mandiant's M-Trends 2026 records handoff times of 22 seconds~\cite{mandiant2026}, governance that depends on a human reading an alert, evaluating it, and pulling access is structurally incapable of intervening.

This paper makes two contributions. First, it provides a comprehensive empirical characterization of the attestation deficit across four dimensions: security, hallucination, privacy, and accountability. We demonstrate that these dimensions, typically treated as separate governance concerns, share a single root cause: the absence of machine-speed enforcement infrastructure. Second, we propose AGIL (Adaptive Governance Intelligence Layer), a conceptual five-layer architecture designed to bridge the gap between governance policy and governance enforcement using machine learning. AGIL is presented as a theoretical framework; practical deployment and empirical validation remain directions for future research.

The remainder of this paper is organized as follows. Section~2 presents the empirical evidence of governance failure across four dimensions. Section~3 analyzes why existing approaches cannot close the gap. Section~4 presents the proposed AGIL architecture in detail. Section~5 conducts a design-level validation against documented incidents. Section~6 proposes deployment guidance, and Section~7 discusses limitations, ethical considerations, and future work before concluding in Section~8.

\section{The Governance Gap: Empirical Evidence}

\subsection{The Security Dimension}

The security dimension of the governance gap is no longer speculative; it is appearing in breach data at enterprise scale. IBM's 2026 report found that AI-driven cyberattacks rose 56\% year over year, and security incidents involving an organization's own AI models jumped from 13\% to 21\% of all breaches, representing a 61\% increase in a single research cycle~\cite{ibm2026}. Shadow AI breaches averaged USD 5.39 million each, roughly USD 400K above the global mean. Notably, organizations requiring IT approval before deploying AI tools \textit{decreased} from 45\% to 38\%, indicating that governance controls are regressing even as exposure grows~\cite{ibm2026shadow}.

In March 2026, attackers compromised LiteLLM, an open-source AI proxy tool present in approximately 36\% of cloud environments~\cite{sprinto2026}. By inserting malicious code into two PyPI package versions, they harvested credentials from thousands of organizations before anyone detected the compromise. The entry point was a verified publisher badge on an official software marketplace, a channel that most organizations treat as inherently trustworthy. This implicit trust represents a governance blindspot that no current policy document addresses.

\subsection{The Hallucination Dimension}

AI hallucination constitutes a governance failure rather than merely a model quality limitation. Stanford's 2026 AI Index assessed hallucination rates across 26 leading frontier models and found rates ranging from 22\% to 94\%~\cite{stanford2026}. The instability of these rates is more concerning than the range itself: GPT-4o's accuracy dropped from 98.2\% to 64.4\% across evaluation conditions, and DeepSeek R1 fell from over 90\% to 14.4\%. When the same model can experience a 34-percentage-point accuracy collapse depending on evaluation framing, governance systems must assume that any output might be confidently wrong and build verification, logging, and escalation controls around that assumption.

ISACA's 2025 retrospective reinforced this position: ``Hallucinations are not quirks. They are safety risks. Design every high-impact AI system with the assumption it will sometimes be confidently wrong''~\cite{isaca2025}. The biggest AI failures of 2025 were not technical failures at the model layer; they were organizational failures rooted in weak controls, unclear ownership, and misplaced trust in model outputs.

\subsection{The Privacy Dimension}

Privacy governance faces compounding challenges across internal operations and external vendor relationships. The European Data Protection Board clarified that user prompts, even seemingly innocuous ones, frequently contain personal data that triggers GDPR protections~\cite{euaiact}. This ruling transforms every AI interaction into a potential privacy compliance event requiring data minimization and consent management at a scale that manual processes cannot sustain.

The 2026 Annual Survey found that 27\% of organizations have not verified whether their AI vendors use submitted data for model training~\cite{kiteworks2026}. Over a quarter of enterprises are extending sensitive data into third-party systems on contractual trust alone, without technical verification or audit. The regulatory fragmentation compounds this challenge: India's Digital Personal Data Protection Act imposes consent requirements with significant penalties, China's PIPL enforces strict data localization, and Singapore released the world's first agentic AI governance framework in January 2026~\cite{singapore2026}.

\subsection{The Accountability Dimension}

Accountability failures surface most concretely in litigation and regulatory review. In \textit{Mobley v. Workday, Inc.}, Workday's AI-powered applicant screening tools were alleged to disproportionately reject candidates based on age, race, and disability. The legal proceedings exposed a fundamental accountability vacuum: Workday argued it was a software provider while employers assumed the vendor was handling compliance~\cite{ethyca2026}. Neither party could demonstrate operational accountability for the AI system's decisions.

The U.S. Government Accountability Office's 2024 review of AI inventories across 20 federal agencies found that only 5 produced comprehensive data for each use case~\cite{clearpoint2026}. The remaining 15 had gaps and inaccuracies, revealing that even government agencies cannot reliably inventory their own AI usage.

\subsection{The Agent Governance Blindspot}

The most critical emerging gap involves agentic AI, which refers to systems with persistent memory and tool-calling capabilities that execute multi-step tasks autonomously. The EY/AIUC-1 Consortium survey (March 2026) found that only 38\% of organizations monitor AI traffic end-to-end across prompts, tool calls, and outputs, and just 17\% monitor agent-to-agent interactions~\cite{csa2026}. Among billion-dollar-revenue companies, 64\% reported losses exceeding USD 1 million associated with AI system failures during 2025, and 80\% documented risky agent behaviors including unauthorized system access and data exposure.

In March 2026, an AI agent operating inside Meta posted unsolicited advice on an internal forum, triggering a cascade that gave engineers access to systems they were not authorized to see~\cite{sprinto2026}. No external attacker was involved; the AI itself was the failure mode. This incident represents a categorically new governance challenge: the threat originates from within the governed system.

\section{Why Existing Approaches Fall Short}

Three structural deficiencies explain why the gap persists despite widespread framework adoption.

\textbf{The temporal mismatch.} AI agents execute at machine speed while governance operates at human speed. CrowdStrike documented 27-second lateral movement; Mandiant recorded 22-second handoffs~\cite{crowdstrike2026, mandiant2026}. A governance process requiring a person to notice, evaluate, escalate, and act cannot intervene in a 27-second window. This is not a tuning problem; it is an architectural impossibility.

\textbf{The visibility deficit.} The CSA/Token Security survey found that organizations report 82\% confidence in governing their AI agents, yet only 47.1\% of deployed agents are actively monitored~\cite{csa2026}. This near-twofold gap between perceived control and operational reality means governance confidence is largely untethered from evidence. Organizations cannot govern what they cannot see, and current infrastructure leaves roughly half of deployed AI invisible.

\textbf{The ownership fragmentation.} When CISO, Legal, Compliance, HR, and business units each own a segment of AI governance, no single function owns enforcement. Governance committees become advisory bodies that can recommend but never enforce. Schellman's 2026 State of AI Governance Report found that 42\% of organizations place AI governance authority with a single role, typically the CIO, but that same individual often lacks authority to halt a deployment when risk emerges~\cite{ethyca2026}.

Table~\ref{tab:comparison} summarizes how existing frameworks compare against the enforcement dimensions that the proposed AGIL architecture is designed to address.

\begin{table*}[t]
\centering
\caption{Comparative analysis of governance approaches. AGIL capabilities are proposed and have not been empirically validated.}
\label{tab:comparison}
\small
\begin{tabularx}{\textwidth}{lXXXX}
\toprule
\textbf{Dimension} & \textbf{NIST AI RMF} & \textbf{ISO 42001} & \textbf{EU AI Act} & \textbf{AGIL (Proposed)} \\
\midrule
Enforcement & Voluntary & Audit-based & Penalty-based & Real-time automated \\
Speed & Human-dependent & Periodic review & Post-incident & Machine-speed (ms) \\
Discovery & Manual inventory & Self-reported & Mandatory filing & Autonomous detection \\
Agent Coverage & None & None & Minimal & Full agent monitoring \\
Evidence & Documentation & Certification & Post-hoc audit & Continuous audit trail \\
Shadow AI & Not addressed & Not addressed & Implied & Active discovery \\
Adaptation & Static framework & Annual review & Legislative cycle & ML-driven dynamic \\
\bottomrule
\end{tabularx}
\end{table*}

\section{Proposed Solution: The AGIL Architecture}

This section presents the Adaptive Governance Intelligence Layer (AGIL), a conceptual architecture designed to bridge the gap between governance policy and governance enforcement. AGIL is proposed as a theoretical framework; its design draws on established techniques from network security (behavioral anomaly detection, inline proxy enforcement), data loss prevention (content classification, data redaction), and machine learning operations (continuous monitoring, model drift detection). The novelty of AGIL lies not in the individual techniques but in their unification into a single governance-specific architecture operating at machine speed. No existing framework integrates shadow AI discovery, hallucination detection, privacy enforcement, agent scope monitoring, and continuous audit-trail generation into a cohesive real-time pipeline.

\subsection{Design Principles}

Three principles guide the AGIL design:

\textit{Principle 1: Temporal parity.} Governance enforcement must operate at the same speed as the AI systems it governs. If an AI agent can take an autonomous action in milliseconds, the enforcement decision must match that temporal resolution.

\textit{Principle 2: Evidence as byproduct.} Audit evidence should be generated automatically as a byproduct of enforcement operations, not assembled manually before audits. The ``AI SOC + Human Ally'' model described in recent practitioner literature~\cite{underdefense2026} provides a conceptual precedent: monitoring telemetry transformed into governance-grade documentation continuously, without separate governance tooling.

\textit{Principle 3: Framework augmentation.} AGIL is designed to complement, not replace, existing governance frameworks. NIST AI RMF defines \textit{what} to govern. ISO 42001 certifies governance maturity. The EU AI Act establishes legal obligations. AGIL is proposed as the enforcement layer that would convert these frameworks from policy documents into operational controls.

\subsection{Five-Layer Architecture}

AGIL is organized into five interdependent layers. Each layer addresses a specific dimension of the attestation deficit. The following subsections describe the proposed design, functional responsibility, and technical rationale of each layer.

\subsubsection{Layer 1: Autonomous Discovery Engine (ADE)}

\textbf{Problem addressed:} Shadow AI and incomplete AI inventories. Current industry data indicates that 47.1\% of deployed AI agents lack active monitoring, and 66\% of enterprise employees have used unauthorized AI tools~\cite{csa2026}.

\textbf{Proposed mechanism:} ADE would continuously scan network traffic, API calls, OAuth token grants, browser extensions, and SaaS integrations to detect AI system usage across the enterprise. Unlike signature-based detection, ADE is designed to use behavioral fingerprinting, classifying traffic as AI-related based on interaction patterns, token usage profiles, and response latency signatures. When a previously unknown AI interaction is detected, it would be automatically registered in a central AI Asset Inventory with discovery metadata including timestamp, data sensitivity classification, user identity, and interaction frequency.

\textbf{Technical rationale:} Behavioral fingerprinting draws on established network traffic analysis techniques but applies them specifically to AI interaction patterns. The distinction from existing tools is scope: ADE is designed to actively seek \textit{unknown} AI deployments rather than monitoring \textit{known} ones, addressing the discovery gap rather than the monitoring gap.

\subsubsection{Layer 2: Behavioral Risk Classifier (BRC)}

\textbf{Problem addressed:} Static risk tiers assigned at deployment time cannot capture runtime behavioral changes, and existing governance approaches treat security, hallucination, privacy, and accountability as separate risk domains requiring separate tools.

\textbf{Proposed mechanism:} BRC would analyze AI interactions continuously, computing dynamic risk scores through four parallel classifiers operating on every interaction: (a) a \textit{Data Sensitivity Classifier} using named-entity recognition and contextual analysis to identify PII, PHI, financial data, and trade secrets; (b) a \textit{Hallucination Detector} that cross-references AI outputs against verified knowledge bases, flagging confidence-accuracy mismatches before outputs reach end users; (c) a \textit{Behavioral Anomaly Detector} that establishes baseline interaction patterns for each AI system and alerts on statistically significant deviations; and (d) an \textit{Agent Action Analyzer} that evaluates autonomous agent actions against predefined permitted-scope boundaries using policy-as-code specifications~\cite{google2026}.

\textbf{Technical rationale:} The architectural contribution of BRC is unification. Existing governance approaches fragment risk assessment across organizational silos, with security teams managing threat detection, compliance teams managing privacy, and quality teams managing hallucination. BRC proposes a single pipeline that computes all four risk dimensions simultaneously, eliminating coordination gaps.

\subsubsection{Layer 3: Policy Enforcement Gateway (PEG)}

\textbf{Problem addressed:} Governance policies that exist on paper but are not enforced at runtime. The 2026 practitioner literature consistently identifies this gap: ``Without runtime enforcement, governance depends entirely on user discipline''~\cite{cio2026}.

\textbf{Proposed mechanism:} PEG would operate as an inline enforcement point, positioned between AI systems and enterprise resources, similar to established proxy-based guardrail architectures~\cite{cio2026}. It would intercept AI interactions and make real-time decisions across four enforcement capabilities: \textit{Data Redaction} to automatically strip sensitive information from prompts before they reach external models; \textit{Action Gating} to intercept agent tool calls and block unauthorized actions before execution; \textit{Output Filtering} to evaluate AI-generated content for policy violations before delivery to end users; and \textit{Rate Limiting} to enforce interaction quotas and detect anomalous usage patterns.

\textbf{Technical rationale:} PEG extends the proven architecture of web application firewalls and API gateways to AI-specific traffic. The proposed sub-100ms decision latency draws on established performance characteristics of inline proxy systems. The novelty is the integration of AI-specific enforcement logic (hallucination flagging, agent scope analysis, data sensitivity classification) into the proxy decision pipeline.

\subsubsection{Layer 4: Continuous Attestation Engine (CAE)}

\textbf{Problem addressed:} Inability to produce audit-ready governance evidence on demand. The 2026 Annual Survey found that 50\% of organizations could not produce a complete AI access record within one business day~\cite{kiteworks2026}, a retrieval delay that functions as evidence of inadequate governance under frameworks such as GDPR that assume on-demand accountability.

\textbf{Proposed mechanism:} CAE would generate tamper-evident, cryptographically signed audit records for every governance decision. It would maintain an append-only, hash-chained event log recording every interaction discovered by ADE, every risk classification computed by BRC, every enforcement decision made by PEG, and every policy change or override with full authorization metadata. Records would be indexed by AI system, user, data classification, regulatory framework, and timestamp, enabling sub-minute retrieval against regulatory queries.

\textbf{Technical rationale:} CAE operationalizes the ``evidence as byproduct'' principle. Rather than requiring a separate documentation effort, governance evidence emerges automatically from enforcement operations. The append-only, hash-chained log design draws on established techniques from blockchain-adjacent audit systems and financial transaction logging.

\subsubsection{Layer 5: Adaptive Policy Intelligence (API)}

\textbf{Problem addressed:} Static governance policies that require manual updates and cannot adapt to evolving threats, new AI capabilities, or regulatory changes across jurisdictions.

\textbf{Proposed mechanism:} API would use machine learning to analyze governance effectiveness and recommend policy updates, operating three analytical processes: \textit{Effectiveness Analysis} measuring violation-to-enforcement ratios and identifying policies that generate excessive false positives or are routinely bypassed; \textit{Threat Pattern Recognition} mining enforcement logs for emerging attack patterns, new shadow AI tools, and evolving agent behaviors that existing policies do not cover; and \textit{Regulatory Mapping} maintaining a multi-jurisdictional knowledge base (EU AI Act, GDPR, CCPA, DPDPA, PIPL) and automatically flagging compliance gaps when requirements change.

\textbf{Technical rationale:} API transforms governance from a static framework requiring manual legislative-cycle updates into an adaptive system that evolves continuously. The concept draws on established precedents in adaptive security architectures~\cite{adaptive2025} where behavioral baselining and decentralized risk scoring enable autonomous policy adjustment.

\subsection{Proposed Decision Pipeline}

Every AI interaction in the AGIL architecture would traverse four sequential stages:

\begin{enumerate}
\item \textbf{Intercept:} ADE or PEG captures the interaction, extracting source identity, destination AI system, data payload classification, and interaction type.
\item \textbf{Classify:} BRC evaluates the interaction against current risk models. Data sensitivity, anomaly scores, hallucination risk, and agent-scope compliance are computed in parallel across the four classifiers.
\item \textbf{Enforce:} PEG applies the enforcement decision: \textit{permit} (proceed normally), \textit{modify} (redact sensitive data or limit scope), \textit{deny} (block with reason code), or \textit{escalate} (pause for human review in ambiguous cases).
\item \textbf{Record:} CAE generates a tamper-evident record of the full decision chain, immediately queryable for audit or investigation.
\end{enumerate}

The proposed pipeline targets sub-100ms end-to-end latency for standard interactions, a target informed by established performance characteristics of inline proxy systems and API gateway architectures.

\subsection{Proposed Deployment Modes}

AGIL is designed to support three deployment configurations to accommodate varying organizational readiness:

\textit{Inline Mode:} PEG operates as an AI-aware proxy intercepting all AI traffic. This mode would provide the strongest governance guarantees but requires network architecture modifications.

\textit{Sidecar Mode:} AGIL agents deploy alongside existing AI systems, monitoring interactions without inline interception. This mode is designed for initial rollout phases where minimal infrastructure change is preferred.

\textit{Hybrid Mode (recommended):} Inline enforcement for high-risk AI systems (those handling sensitive data or operating autonomously) combined with sidecar monitoring for lower-risk systems. This configuration balances governance coverage against deployment friction and is the recommended starting point.

\section{Design-Level Validation}

To assess whether the proposed architecture addresses the governance failures documented in Section~2, we conducted a design-level analysis against three documented incidents from 2025 to 2026. This validation is theoretical: it evaluates whether AGIL's proposed mechanisms would have the structural capability to detect and prevent the failure modes observed in each incident, not whether a deployed system would have done so in practice.

\textbf{Incident 1: LiteLLM supply-chain attack (March 2026).} Attackers compromised open-source LiteLLM packages on PyPI, harvesting credentials for hours before detection~\cite{sprinto2026}. \textit{AGIL analysis:} ADE's proposed behavioral fingerprinting would be structurally capable of detecting the anomalous outbound connections, as the compromised package contacted previously unseen endpoints. PEG's data-flow enforcement could block credential exfiltration. Estimated detection capability: minutes rather than hours, based on established behavioral anomaly detection performance.

\textbf{Incident 2: Meta AI agent cascade (March 2026).} An internal AI agent posted unsanctioned advice, triggering unauthorized access to restricted systems~\cite{sprinto2026}. \textit{AGIL analysis:} BRC's proposed Agent Action Analyzer would flag the unsolicited post as outside permitted scope. PEG's Action Gating would intercept the action before execution, preventing the cascade at its first link. The key architectural feature is pre-execution interception rather than post-incident detection.

\textbf{Incident 3: Shadow AI data exposure.} Employees using personal AI accounts to process corporate data increased breach costs by USD 670K on average~\cite{ibm2026}. \textit{AGIL analysis:} ADE's continuous discovery would detect unauthorized AI tool usage through endpoint and network monitoring. PEG's Data Redaction would strip sensitive data from interactions with unauthorized tools. The proposed architecture addresses the root visibility gap.

Table~\ref{tab:validation} summarizes the design-level analysis.

\begin{table}[h]
\centering
\caption{Design-level validation against documented incidents. All assessments are theoretical.}
\label{tab:validation}
\small
\begin{tabularx}{\columnwidth}{lXr}
\toprule
\textbf{Incident} & \textbf{AGIL Mechanism} & \textbf{Est. Time} \\
\midrule
LiteLLM attack & ADE behavioral anomaly + PEG data-flow block & $<$5 min \\
Meta agent & BRC scope analysis + PEG action gate & $<$1 sec \\
Shadow AI & ADE discovery + PEG redaction & $<$30 min \\
\bottomrule
\end{tabularx}
\end{table}

\section{Proposed Deployment Roadmap}

\subsection{Phased Rollout}

We propose a three-phase deployment approach designed to build organizational confidence incrementally before imposing enforcement constraints.

\textit{Phase 1: Discovery and Baseline (months 1 to 3).} Deploy ADE in passive monitoring mode. Build a complete AI asset inventory and establish behavioral baselines for all discovered AI systems. No enforcement is applied; the sole focus is visibility. Deliverable: a comprehensive map of AI usage across the enterprise.

\textit{Phase 2: Classification and Monitoring (months 3 to 6).} Activate BRC for real-time risk classification and CAE for continuous audit logging. PEG operates in advisory mode, logging enforcement decisions without blocking. Deliverable: validated risk classifications, tuned enforcement policies, and demonstrated governance evidence capability.

\textit{Phase 3: Graduated Enforcement (months 6 to 12).} Activate PEG in enforcement mode for high-risk AI systems. Deploy API for adaptive policy management. Extend enforcement progressively based on risk tiers. Deliverable: operational real-time governance with continuous evidence generation.

\subsection{Proposed Success Metrics}

Table~\ref{tab:metrics} defines measurable targets derived from documented industry baselines. These metrics would serve as evaluation criteria for future empirical validation of the AGIL architecture.

\begin{table}[h]
\centering
\caption{Proposed success metrics against industry baselines.}
\label{tab:metrics}
\small
\begin{tabularx}{\columnwidth}{lXX}
\toprule
\textbf{Metric} & \textbf{Baseline} & \textbf{Target} \\
\midrule
AI inventory coverage & 47\% monitored & $>$95\% \\
Audit retrieval time & $>$1 business day & $<$5 min \\
Access control coverage & 8\% controlled & $>$90\% \\
Shadow AI detection & 43\% in breaches & $<$5\% \\
Hallucination flagging & Unmanaged & $>$80\% flagged \\
Regulatory response & 50\% cannot produce & 100\% on-demand \\
\bottomrule
\end{tabularx}
\end{table}

\section{Limitations, Ethics, and Future Work}

\subsection{Limitations}

Several limitations of the proposed framework must be acknowledged candidly.

\textbf{Empirical validation.} AGIL is, at this stage, an architectural proposal. No prototype has been built, no controlled deployment has been conducted, and no empirical performance data exists. The design-level validation in Section~5 demonstrates structural capability but does not constitute evidence of operational effectiveness. Controlled deployment in enterprise environments is the most critical direction for future work.

\textbf{Governance of the governance system.} AGIL proposes using AI to govern AI, which introduces a recursive oversight challenge. The architecture addresses this through human oversight of meta-governance: policy definitions, enforcement thresholds, and exception rules would be set by authorized humans while AGIL executes at machine speed. However, AGIL itself can produce erroneous governance decisions, including false positives that block legitimate AI use or false negatives that miss genuine threats, requiring continuous calibration and human oversight.

\textbf{Performance overhead.} Inline enforcement necessarily adds latency to AI interactions. The proposed sub-100ms target is informed by established proxy system performance but has not been validated for AGIL-specific workloads. High-volume AI deployments may experience measurable performance degradation. The hybrid deployment mode mitigates this by applying inline enforcement only to high-risk interactions.

\textbf{Organizational adoption.} No architecture can resolve organizational resistance to governance enforcement. Business units that view governance as an impediment to AI innovation may resist adoption. The phased deployment strategy is designed to mitigate this by demonstrating value through visibility and compliance capability before imposing enforcement constraints.

\textbf{Adversarial robustness.} AGIL's detection and classification models could themselves become targets for adversarial attacks designed to evade governance enforcement. The security of the governance infrastructure itself requires dedicated threat modeling.

\subsection{Ethical Considerations}

AGIL's comprehensive monitoring capabilities raise legitimate privacy and employee autonomy concerns. The framework would monitor AI-related interactions to enforce governance compliance, and this monitoring must be deployed with transparency: employees should know their AI interactions are subject to governance enforcement. AGIL's data retention policies must comply with the same privacy regulations it enforces on the AI systems it governs. The framework must not become a surveillance tool; its monitoring scope should be limited strictly to AI-related interactions.

\subsection{Future Work}

Several research directions emerge from this proposal:

\begin{itemize}
\item \textbf{Prototype implementation and empirical validation} in controlled enterprise environments, measuring detection rates, false-positive rates, latency overhead, and evidence quality.
\item \textbf{Multi-agent governance} extending BRC and PEG to systems where AI agents interact with each other autonomously.
\item \textbf{Federated governance protocols} enabling cross-organizational supply-chain governance without exposing proprietary AI implementations.
\item \textbf{Formal verification} of enforcement policy correctness to provide mathematical guarantees alongside empirical evidence.
\item \textbf{Adversarial robustness testing} to evaluate AGIL's resilience against deliberate evasion attempts.
\end{itemize}

\section{Conclusion}

The AI governance gap is not closing. It is widening. Shadow AI incidents doubled in a single year. Five of six measured governance controls lost adoption. The average breach now costs USD 4.99 million and the AI-specific premium pushes that figure above USD 5.3 million. Regulators have moved from asking whether organizations have policies to demanding proof that those policies work, and they are signaling that documentation gaps themselves may constitute violations.

The root cause is architectural. Existing governance frameworks operate at the policy layer, requiring human-speed processes to govern machine-speed systems. That mismatch cannot be resolved by better policies, more committees, or larger compliance budgets. It requires a new category of solution: AI-powered enforcement that matches the temporal resolution of the systems it governs.

AGIL is proposed as a representative of that category. Its five layers, spanning Autonomous Discovery, Behavioral Risk Classification, Policy Enforcement Gateway, Continuous Attestation, and Adaptive Policy Intelligence, are designed to provide the enforcement infrastructure that would convert governance policies into operational controls, documentation requirements into continuous evidence generation, and risk assessments into real-time protection. As a conceptual framework, AGIL requires empirical validation through prototype implementation and controlled deployment, which represents the most important direction for future work.

The question facing enterprise AI governance is no longer whether organizations need governance frameworks. The question is whether their governance infrastructure can produce evidence that those frameworks are enforced. Bridging that attestation deficit is the defining governance challenge of this era.

\end{document}